\documentclass[journal]{IEEEtran}

\usepackage{xcolor,soul,framed} %,caption

\colorlet{shadecolor}{yellow}
\usepackage[pdftex]{graphicx}
\graphicspath{{../pdf/}{../jpeg/}}
\DeclareGraphicsExtensions{.pdf,.jpeg,.png}

\usepackage[cmex10]{amsmath}
\usepackage{array}
\usepackage{mdwmath}
\usepackage{mdwtab}
\usepackage{eqparbox}
\usepackage{url}

\usepackage{times}
\usepackage{helvet}
\usepackage{courier}
\usepackage{hyperref}       % hyperlinks
\usepackage{url}   
\usepackage{xcolor}
\usepackage{soul}
\usepackage[utf8]{inputenc}
\usepackage[small]{caption}
\usepackage[cmex10]{amsmath}
\usepackage{epsfig,amsfonts,amssymb,times,subfigure,floatflt,wrapfig,enumerate,tikz,url,bm}
\usepackage{algorithmicx,algpseudocode}

\usepackage{booktabs}
\usepackage{siunitx}
\usepackage{tabu}
\usepackage{multirow}
\newtheorem{example}{Example}

\newtheorem{definition}{Definition}

\newtheorem{alg}{Algorithm}
\newcommand{\mat}[1]{\bm{#1}}
\newcommand{\ten}[1]{\bm{\mathcal{#1}}}

\begin{document}
\bstctlcite{IEEEexample:BSTcontrol}
    \title{Deep Compression of Sum-Product Networks on Tensor Networks}
  \author{Ching-Yun Ko\textsuperscript{*},
      Cong Chen\textsuperscript{*},
      Yuke Zhang,
      Kim Batselier
      and~Ngai Wong% <-this % stops a space
      \\
chencong@eee.hku.hk,
cyko@eee.hku.hk,
yukezhan@usc.edu,
k.batselier@tudelft.nl,
nwong@eee.hku.hk

  %\thanks{Manuscript received July 10, 2012. This paper is an expanded paper from the IEEE MTT-S Int. Microwave Symposium held on June 17-22, 2012 in Montreal, Canada. This work was funded in part by the Office of Naval Research under the Defense Advanced Research Projects Agency (DARPA) Microscale Power Conversion (MPC) Program under Grant N00014-11-1-0931, and in part by the Advanced Research Projects Agency-Energy (ARPA-E), U.S. Department of Energy, under Award Number DE-AR0000216.}
  \thanks{\textsuperscript{*}~C.Y. Ko and C. Chen equally contributed to this work.}% <-this % stops a space
  %\thanks{C.Y. Ko, C. Chen and N. Wong are with the Department of Electrical and Electronic Engineering, The University of Hong Kong (e-mail: cyko@eee.hku.hk; chencong@eee.hku.hk; nwong@eee.hku.hk).}%
  %\thanks{Y. Zhang is with the Department of}
  
  %\thanks{K. Batselier is with the Department of (e-mail: k.batselier@tudelft.nl).} 
  }

%\markboth{IEEE TRANSACTIONS ON NEURAL NETWORKS AND LEARNING SYSTEMS
%}{Roberg \MakeLowercase{\textit{et al.}}: High-Efficiency Diode and Transistor Rectifiers}

% ====================================================================
\maketitle

% === ABSTRACT ====================================================================
% =================================================================================
\begin{abstract}
Sum-product networks (SPNs) represent an emerging class of neural networks with clear probabilistic semantics and superior inference speed over graphical models. This work reveals a strikingly intimate connection between SPNs and tensor networks, thus leading to a highly efficient representation that we call tensor SPNs (tSPNs). For the first time, through mapping an SPN onto a tSPN and employing novel optimization techniques, we demonstrate remarkable parameter compression with negligible loss in accuracy.
\end{abstract}

% === KEYWORDS ====================================================================
% =================================================================================

\IEEEpeerreviewmaketitle

% ====================================================================
% ====================================================================
% ====================================================================

% === I. INTRODUCTION =============================================================
% =================================================================================

\section{Introduction}
\label{sec:intro}
\IEEEPARstart{S}{ince} the inception of sum-product networks (SPNs)~\cite{Poon2011}, a multitude of works have emerged with respect to their structure and weight learning, e.g.,~\cite{Vergari2015,Zhao2016,Butz2017}, as well as their application in image completion, speech modeling, semantic mapping and robotics, e.g.,~\cite{Zheng2018}, just to name a few. An SPN exhibits a clear semantics of mixtures (sum nodes) and features (product nodes). 

In short, given a high-dimensional dataset $\mat{x}_k\in\mathbb{R}^d$ ($k=1\ldots N$), an SPN learns and encodes a probability distribution over the data and implicit latent (hidden) variables. Compared to other probabilistic graphical models like Bayesian and Markov networks with \#P or NP-hard computation, an SPN enjoys a tractable exact inference cost, and its learning is relatively simple and fast. 

On the other hand, there has been an exploding number of works on tensors (a multilinear operator rooted in physics)~\cite{Kolda2009} including their connection and utilization in various engineering fields such as signal processing~\cite{Cichocki2015}, and lately also in neural networks and machine learning~\cite{Cohen2017,Khrulkov2018,Chen2018}. The power of tensors lies in their ability to lift the curse of dimensionality, reducing computational and storage complexity from exponential to linear cost. 

In this work, we establish a natural tensor representation of an SPN which we call a tensor SPN (tSPN). By utilizing the tensor notion, we essentially expand the set of tractable SPNs to match that of compact and deep SPNs. We focus on structure compression rather than the learning of SPNs. Indeed, our work leverages on the wealth of existing SPN learning algorithms such as~\cite{Gens2013,Vergari2015,Zhao2016}, and attempts to turn their inherently wide SPN tree outputs (owing to the intrinsic way of learning through partitioning the data matrix) into a ``deep'' tree via tensor network techniques subject to a non-negativity constraint. This is in-line with the recent trend on pruning neural networks (NNs)~\cite{Han2015}, but operates quite differently in the regime of tensor networks with the imposition of a non-negative constraint not needed in the NN counterparts.

 In particular, sparsifying an SPN and transforming it into a tSPN brings about the following advantages:
\begin{itemize}
\item It automatically enables sharing of weights through the tensor network cores, often resulting in significant parameter compression.
\item Fewer parameters can facilitate faster computation, lower power consumption, and less stringent cooling requirements in hardware implementations.
\item The above is crucial in downstream edge computing, which translates into better user privacy, lower reliance on the cloud and lower wireless bandwidth for over-the-air network update on edge devices.
%\item Implicit regularization due to the underlying tensor structure, which effectively prevents overfitting. 
\end{itemize}
To our knowledge, this paper is the first to systematically transform an SPN into a tensor format, specifically, a tensor network counterpart named tSPN. The following sections cover SPN and tensor basics, and then introduce the tSPN conversion. Numerical experiments then demonstrate the remarkable compression in the number of network parameters, with negligible loss in the probabilistic modeling accuracy.

\section{Preliminaries}
\subsection{SPN basics}
\label{subsec:spn}
\begin{figure}[t] 
\begin{center} 
\includegraphics[width=0.35\textwidth]{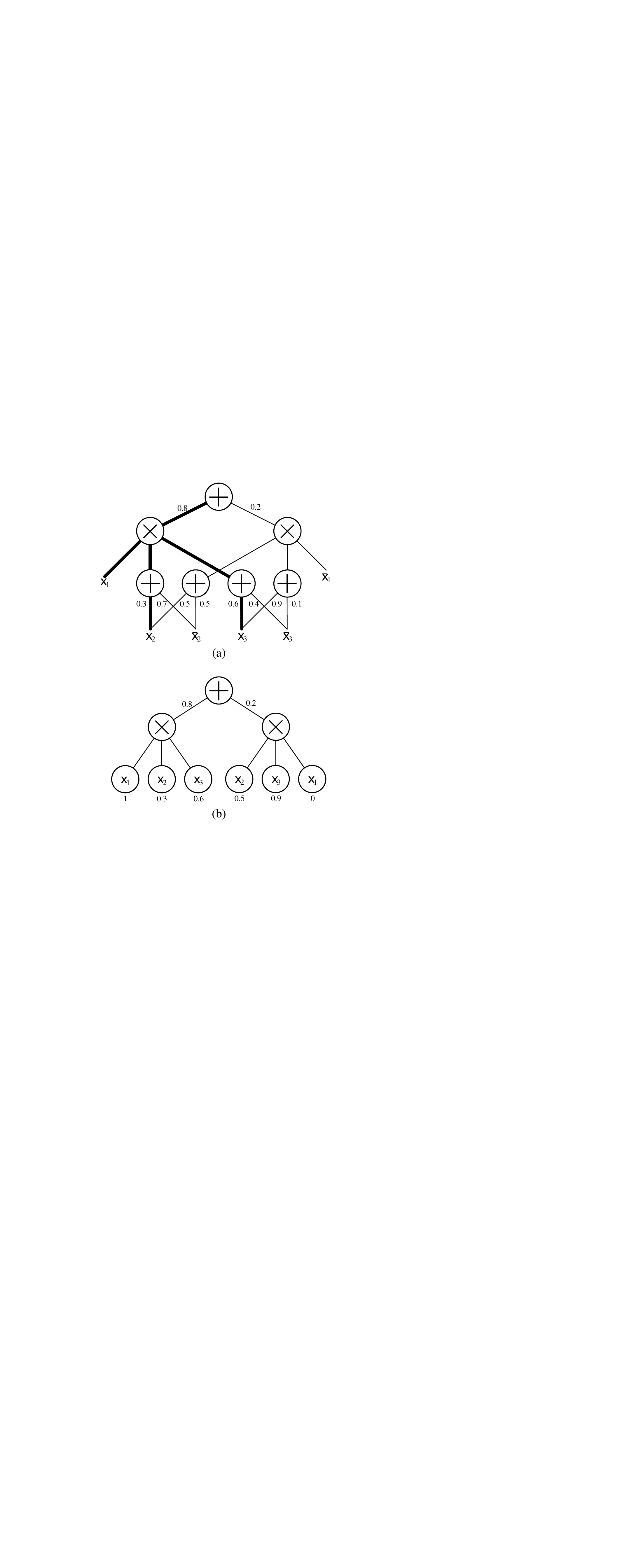}
\caption{(a) An example SPN with boolean variables and (b) its equivalent leaf-node representation where, e.g., the lower left $x_2$ ball represents the Bernoulli function $0.3x_2+0.7\bar{x}_2$. The bold edges in (a) denote an induced tree example (see Definition~\ref{def:indtree}).}
\label{fig:spn_ex}
\end{center}
\end{figure}
\begin{figure}[t] 
\begin{center} 
\includegraphics[width=0.35\textwidth]{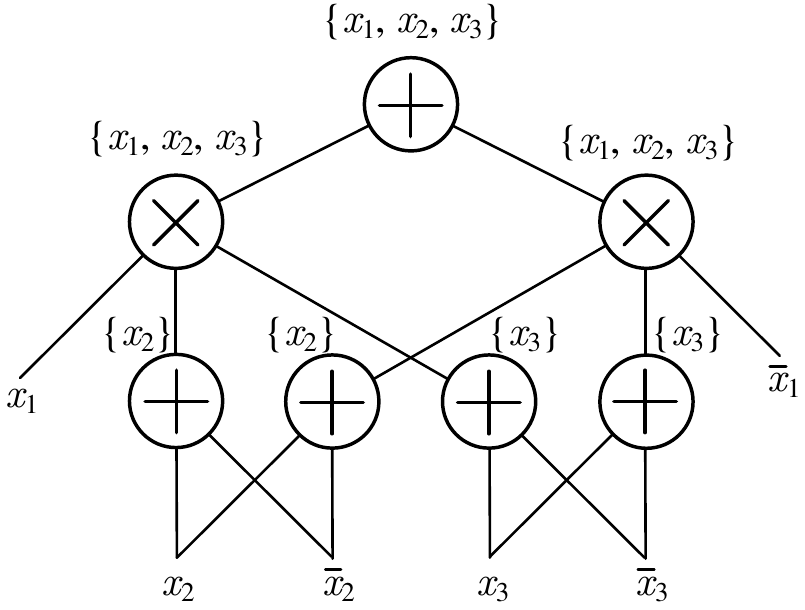}
\caption{The scopes, denoted by $\{\circ\}$, for every node in the example SPN.}
\label{fig:scope}
\end{center}
\end{figure}
%\begin{figure}[t] 
%\begin{center} 
%\includegraphics[width=0.2\textwidth]{figures/indtree.pdf}
%\caption{A particular induced tree of the example SPN.}
%\label{fig:indtree}
%\end{center}
%\end{figure}

We use a slightly modified SPN example from~\cite{Poon2011,Gens2012}, shown in Fig.~\ref{fig:spn_ex}(a), to motivate some important concepts and operations of SPNs. Boolean variables are chosen for the ease of illustration, but their generalization to multi-nominal or continuous variables are straightforward~\cite{Poon2011}. To begin with, an SPN is a directed acyclic graph with alternating layers of sum and product (internal) nodes and a root node on top. The edges emanating downwards from sum nodes have non-negative weights, while the edges emanating downwards from product nodes are all of unit weight. The leaves contain the set of random variables $\mat{X}=\{X_1,\ldots,X_d\}$. For boolean variables, the indicator functions $[X_i]$ and $[\bar{X}_i]$ are 1 when $X_i$ and $\bar{X}_i$ are 1, respectively, and 0 otherwise. We adopt the abbreviations $x_i$ and $\bar{x}_i$ for $[X_i]$ and $[\bar{X}_i]$, respectively.
% Consequently, the leaves in the SPN in Fig.~\ref{fig:spn_ex} are Bernoulli distributions $f_i$'s with probability $p_i$'s such that $f_i(x_i,\bar{x}_i)=p_i x_i+(1-p_i)\bar{x}_i$.

\begin{definition}
The \emph{scope} of an SPN is the set of variables appearing in its leaves. The scope of an internal sum or product node is the scope of the corresponding \emph{sub-SPN} rooted at that node, as illustrated in Fig.~\ref{fig:scope}.%
\end{definition}
\begin{definition}
An SPN is called \emph{complete} when all children of a sum node have identical scope. It is called \emph{decomposable} when all children of a product node have disjoint scopes. An SPN is called \emph{valid} when all its sum nodes are complete and all its product nodes are decomposable. The sum nodes have the semantics of a \emph{mixture} of components, while product nodes represent \emph{features}. An SPN is called a \emph{normalized} SPN when the edges emanating from a sum node have a total weight of one, and is an unnormalized SPN otherwise.%
\end{definition}
Consequently, the SPN in Fig.~\ref{fig:spn_ex}(a) is a valid and normalized SPN. We use $\mathcal{S}_{\mat{w}} (\mat{x})$ to denote the value of an SPN where $\mat{w}$ is a vector containing all (non-negative) weights in the network, and $\mat{x}\in\mathbb{R}^d$ is the vector of all random variables. A distribution is called \emph{tractable} if any marginal probability can be computed in linear time.
\begin{definition}
\label{def:indtree}
An induced tree~\cite{Zhao2016} is a sub-tree of an SPN originating from the root following two rules: i) only one edge out of a sum node is selected at a time; ii) all edges out of a product node are selected. It can be readily checked that the total number of induced trees arising from an SPN can be computed by $\tau=\mathcal{S}_{\mat{1}} (\mat{1})$, i.e., by setting $\mat{w}=\mat{1}$ and $\mat{x}=\mat{1}$ where $\mat{1}$ is the all-ones vector.%
\end{definition}
For instance, the bolded edges in Fig.~\ref{fig:spn_ex}(a) denotes an induced tree by selecting the left route out of each sum node. The notion of a network polynomial~\cite{Darwiche2003} comes in handy at this point.
\begin{definition}
Let $f(\mat{x})$ be the probability mass function of a set of discrete random variables $\mat{X}=\{X_1,\ldots,X_d\}$. The network polynomial of $f(\mat{x})$ is the multilinear polynomial $\sum_{\mat{x}} \prod_{\mat{x}} f(\mat{x}) \prod_{\mat{x}} \lambda(\mat{x})$, where $\prod_{\mat{x}} \lambda(\mat{x})$ is the product of evidence indicators that has a value of 1 in the state $\mat{x}$.
% A network polynomial is a multivariate polynomial of the indicator variables where each variable has degree one, or in other words, multilinear in its variables. A network polynomial contains an exponential number of terms, one for each possible state of $\mat{x}$.
\end{definition}
Any joint probability function of $d$ $I$-valued discrete random variables is represented by $I^d$ probabilities. The corresponding network polynomial has therefore $I^d$ terms. For example, the joint probability function of the SPN in Fig.~\ref{fig:spn_ex} has a network polynomial that consists of $2^3=8$ terms:%
\begin{align}
\label{eqn:np}
&f(\mat{x})=\mathcal{S}_{\mat{w}}(\mat{x})\nonumber\\
&=(0.8)(0.3)(0.6)x_1 x_2 x_3+(0.8)(0.3)(0.4)x_1 x_2 \bar{x}_3\nonumber\\
&+(0.8)(0.7)(0.6)x_1 \bar{x}_2 x_3+(0.8)(0.7)(0.4)x_1 \bar{x}_2 \bar{x}_3\nonumber\\
&+(0.2)(0.5)(0.9)\bar{x}_1 x_2 x_3+(0.2)(0.5)(0.1)\bar{x}_1 x_2 \bar{x}_3\nonumber\\
&+(0.2)(0.5)(0.9)\bar{x}_1 \bar{x}_2 x_3+(0.2)(0.5)(0.1)\bar{x}_1 \bar{x}_2 \bar{x}_3.%
\end{align}%
Therefore an equivalence is drawn between an SPN and its network polynomial representation, which in turn encodes a probability function.

The beauty of an SPN lies in its exact and tractable inference. Equation~(\ref{eqn:np}) is an instance of a normalized SPN. For an unnormalized SPN, there are two ways to build its probability function. One is to scale the edge weights out of each sum node such that they add up to one, i.e., turning it back into a normalized SPN. Alternatively, we can compute the \emph{partition function} in one bottom-up pass by setting $\mat{x}=\mat{1}$:
\begin{align*}
% \label{eqn:Z}
Z&=\mathcal{S}_{\mat{w}}(\mat{1}) 
\end{align*}
such that $\mathcal{P}(\mat{x})=\mathcal{S}_{\mat{w}}(\mat{x})/Z$ is a probability function.
\begin{example}
Assuming a normalized SPN, the probability of a fully specified state (also called a complete evidence) $\mat{x}$, e.g., $x_1=1$, $x_2=0$, $x_3=1$ in Fig.~\ref{fig:spn_ex}(a), is easily computed through a bottom-up pass by setting $x_i=1$ and $\bar{x}_i=0$ for $i=1,3$ and $x_2=0$ and $\bar{x}_2=1$. 
\end{example}
\begin{example}
Assuming a normalized SPN, the probability of an evidence, e.g., $x_1=1$ in Fig.~\ref{fig:spn_ex}(a), can be computed by marginalizing over $x_2$ and $x_3$. This is computed through a bottom-up pass by setting $x_1=1$ and $\bar{x}_1=0$, and $x_i=1$ and $\bar{x}_i=1$ for $i=2,3$.
\end{example}
The above two examples are readily verified by comparing with~(\ref{eqn:np}). Similar operations allow us to compute the conditional probability and most probable explanation (MPE) by replacing sum nodes with max nodes, called a max-product network, in a similarly efficient manner~\cite{Peharz2016}.

To articulate with tensor-based reformulation of the SPN, we will need to slightly modify the induced tree (Definition~\ref{def:indtree}) by terminating at the leaf nodes, i.e., the bottom nodes of the SPN in Fig.~\ref{fig:spn_ex}(b) instead of down to individual $x_i$ or $\bar{x}_i$.  With reference to Fig.~\ref{fig:spn_ex}(b), the SPN contains two such induced trees corresponding to the left and right branches originating from the root sum node.

\begin{figure}[t] 
\begin{center} 
\includegraphics[width=0.4\textwidth]{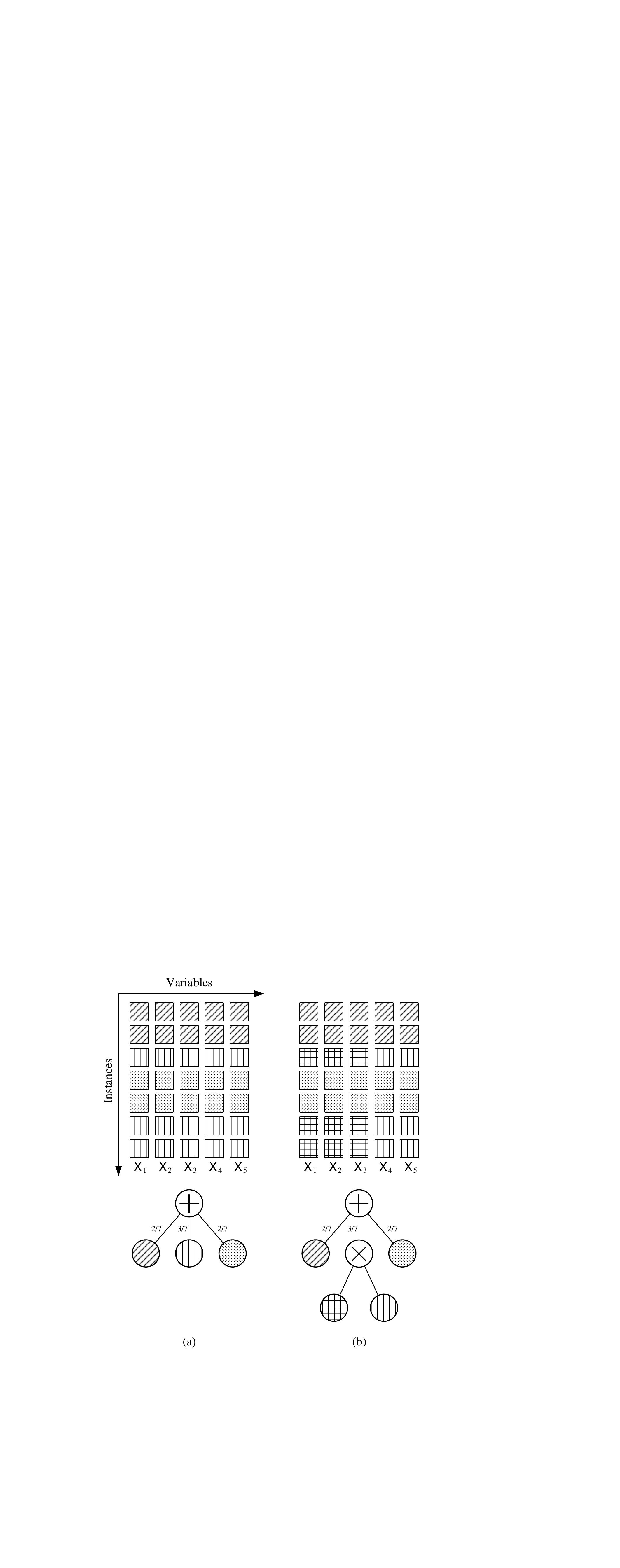}
\caption{The learnSPN operations: (a) slicing (b) chopping.}
\label{fig:learnSPN}
\end{center}
\end{figure}
In fact, prevailing SPN learning algorithms or alike, e.g, LearnSPN, SPN-B and SPN-BT~\cite{Gens2013,Vergari2015}, all produce SPN trees terminating at leaf nodes. Although SPN illustrations often utilize networks with shared weights (e.g., the two top branches in Fig.~\ref{fig:spn_ex} are shared among many induced trees), conventional learning algorithms are all based on the ``slice'' and ``chop'' operations on the dataset matrix~\cite{Butz2017}, or their variants with additional regularization constraints. A toy example illustrates the basic learnSPN flow. Referring to Fig.~\ref{fig:learnSPN}(a), the slicing operation constructs children of a sum node by clustering similar sample instances. This is often done via k-means or expectation-maximization (EM) for Gaussian mixture models (GMMs). In Fig.~\ref{fig:learnSPN}(b), the chopping operation constructs children of a product node by grouping dependent variables. This is often done by the G-test or mutual information methods wherein a scoring formula is used to determine variables belonging to the same group, if any. These so-called hierarchical divisive clustering steps are surprisingly simple and effective, but they never look back and this leads to inherently wide SPN tree structures. For example, in the standard NLTCS benchmark, learnSPN (with default hyperparameters) generates an SPN with 19 layers and 1420 leaf nodes, even though there are only 16 variables. This demonstrates that existing learning algorithms do not readily produce shared edges (and weights) across different induced trees, and do not generate SPNs that can otherwise be represented compactly.

\subsection{Tensor basics}
\label{subsec:tensor}
In this article, tensors are high-dimensional arrays that generalize vectors and matrices to higher orders. A $d$-way or $d$-order tensor $\ten{A} \in\mathbb{R}^{I_1\times I_2\times \cdots\times I_d}$ is an array where each entry is indexed by $d$ indices $i_1,i_2,\ldots,i_d$. We use the convention $1\leq i_k\leq I_k$ for $k=1,\ldots,d$. When $I_1=\ldots =I_d=I$ the tensor is called \emph{cubical}. MATLAB notation is used to denote entries of tensors. Boldface capital calligraphic letters $\ten{A},\ten{B},\ldots$ are used to denote tensors, boldface capital letters $\mat{A},\mat{B},\ldots$ denote matrices, boldface letters $\mat{a},\mat{b},\ldots$ denote vectors, and Roman letters $a,b,\ldots$ denote scalars. A set of $d$ tensors, like that of a tensor train (TT), is denoted as $\ten{A}^{(1)},\ten{A}^{(2)},\ldots,\ten{A}^{(d)}$. The notion of a rank-1 matrix is generalized to tensors in the following manner.
\begin{definition}~\cite[p.~460]{Kolda2009}
For a given set of vectors $\mat{a}_1 \in \mathbb{R}^{N_1},\ldots,\mat{a}_d \in \mathbb{R}^{N_d}$, the corresponding rank-1 tensor $\ten{A} \in \mathbb{R}^{N_1 \times \cdots \times N_d}$ is defined such that
\begin{align*}
\ten{A}(i_1,i_2,\ldots,i_d) := \mat{a}_1(i_1) \mat{a}_2(i_2) \cdots \mat{a}_{d}(i_d).
\end{align*}
 Alternatively, this is written as
\begin{align*}
\ten{A} := \mat{a}_1 \circ \mat{a}_2 \circ \cdots \circ \mat{a}_d,
\end{align*}
where $\circ$ denotes the outer product.
\end{definition}
A rank-$r$ tensor is then defined as the sum of $r$ rank-1 tensors. The generalization of the matrix-vector multiplication to tensors involves a multiplication of a vector with a $d$-way tensor along one of its $d$ modes.
\begin{definition}~\cite[p.~458]{Kolda2009})
The $k$-mode product of a tensor~\mbox{$\ten{A}\in\mathbb{R}^{I_1\times\cdots \times I_d}$} with a vector \mbox{$\mat{u}\in\mathbb{R}^{I_k\times1}$} is denoted \mbox{$\ten{B}=\ten{A}\, {\times_k}\, \mat{u}^T$} and defined by%
\begin{align*}
\nonumber \ten{B}(i_1,\cdots,i_{k-1},i_{k+1}, \cdots, i_d) &= \hfill \\
\sum\limits_{i_k=1}^{I_k}  \mat{u}(i_k) \ten{A}(i_1,\cdots,i_{k-1},i_k,&i_{k+1},\cdots,i_d),
%\label{eqn:kmode}
\end{align*}
where $\ten{B}\in\mathbb{R}^{I_1\times\cdots \times I_{k-1} \times I_{k+1}\times\cdots\times I_d}$.
\end{definition}
The proposed method also requires the knowledge of the matrix Khatri-Rao product as defined below.
\begin{definition}
If $\mat{A} \in \mathbb{R}^{N_1 \times M}$ and~$\mat{C} \in \mathbb{R}^{N_2 \times M}$, then their Khatri-Rao product $\mat{A} \odot \mat{C}$ is the $N_1N_2 \times M$ matrix
\begin{align*}
\begin{pmatrix}
\mat{A}(:,1)\otimes \mat{C}(:,1) & \cdots & \mat{A}(:,M)\otimes \mat{C}(:,M) \\
\end{pmatrix},
%\label{def:khatri}
\end{align*}
where $\otimes$ denotes the standard Kronecker product.
\end{definition}
The storage of a $d$-way tensor with dimensions $N$ requires $N^d$ elements. Tensor decompositions are crucial in reducing the exponential storage requirement of a given tensor. In this article we will make use of the tensor train (TT) decomposition~\cite{oseledets2011tensor}, which is a particular kind of tensor network.
\begin{definition}~\cite[p.~2296]{oseledets2011tensor}\label{def:TT}
A tensor train (TT) representation of a tensor $\ten{A}$ is a set of $d$ 3-way tensors $\ten{A}^{(1)} \in \mathbb{R}^{R_1 \times I_1 \times R_2},\ten{A}^{(2)} \in \mathbb{R}^{R_2 \times I_2 \times R_3},\ldots,\ten{A}^{(d)} \in \mathbb{R}^{R_{d} \times I_d \times R_1}$ such that $\ten{A}(i_1,i_2,\ldots,i_d)$ can be computed from
\begin{align*}
\sum_{r_1,\ldots,r_d}^{R_1,\ldots,R_d} \ten{A}^{(1)}(r_1,i_1,r_2) \ten{A}^{(2)}(r_2,i_2,r_3)\cdots \ten{A}^{(d)}(r_d,i_d,r_1).
\end{align*}
The dimensions $R_1,R_2,\ldots,R_d$ are called the TT-ranks, and the three-way tensors $\ten{A}^{(1)},\ten{A}^{(2)},\ldots,\ten{A}^{(d)}$ are called the TT-cores.
\end{definition}
When $R_1 > 1$, the TT is said to have periodic boundary conditions and is called a tensor loop or tensor ring. In this article, we will always set $R_1=1$, i.e., a standard TT structure.
\textbf{The key idea now is to represent the network polynomial of any joint probability function by a low-rank TT.} In this way, all $I^d$ (if $I_1=\ldots= I_d=I$)
% $k^d$
probabilities can be computed from $\approx O(dIR^2)$ 
% $\approx O(dkR^2)$ 
numbers, where $R$ is the maximal TT-rank. Without loss of generality, we consider only boolean variables $I=2$ from now on.
\begin{definition}\label{def:tspncontract}
For a given network polynomial $f(\mat{x})$ of $d$ binary random variables we define the corresponding TT consisting of $d$ 3-way tensors $\ten{F}^{(1)} \in \mathbb{R}^{1 \times 2 \times R_2},\ten{F}^{(2)} \in \mathbb{R}^{R_2 \times 2 \times R_3},\ldots,\ten{F}^{(d)} \in \mathbb{R}^{R_d \times 2 \times 1}$ such that the evaluation of $f(\mat{x})$ for a given state $\mat{x}$ can be computed from
{
% \footnotesize
\scriptsize
\begin{align*}
\left(\ten{F}^{(1)} \times_2 \begin{pmatrix} x_1 \\ \bar{x}_1 \end{pmatrix}^T\right) \left(\ten{F}^{(2)} \times_2 \begin{pmatrix} x_2 \\ \bar{x}_2\end{pmatrix}^T\right)\cdots \left(\ten{F}^{(d)} \times_2 \begin{pmatrix} x_d \\ \bar{x}_d \end{pmatrix}^T\right).
\end{align*}
}
\end{definition}
Note that the $\ten{F}^{(1)} \times_2 \begin{pmatrix} x_1 & \bar{x}_1 \end{pmatrix}$ and $\ten{F}^{(d)} \times_2 \begin{pmatrix} x_d & \bar{x}_d \end{pmatrix}$ factors are a column and row vector, respectively. The other factors are matrices, such that the whole product results in the scalar $f(\mat{x})$. Small TT-ranks for $f(\mat{x})$ then imply small matrix factors, and a massive reduction in the required number of parameters can be achieved.
%normalized tSPN, dropping zero slices.

\section{Putting an SPN on a tSPN}
\label{sec:spn2tspn}
\begin{figure}[t] 
\begin{center} 
\includegraphics[width=0.45\textwidth]{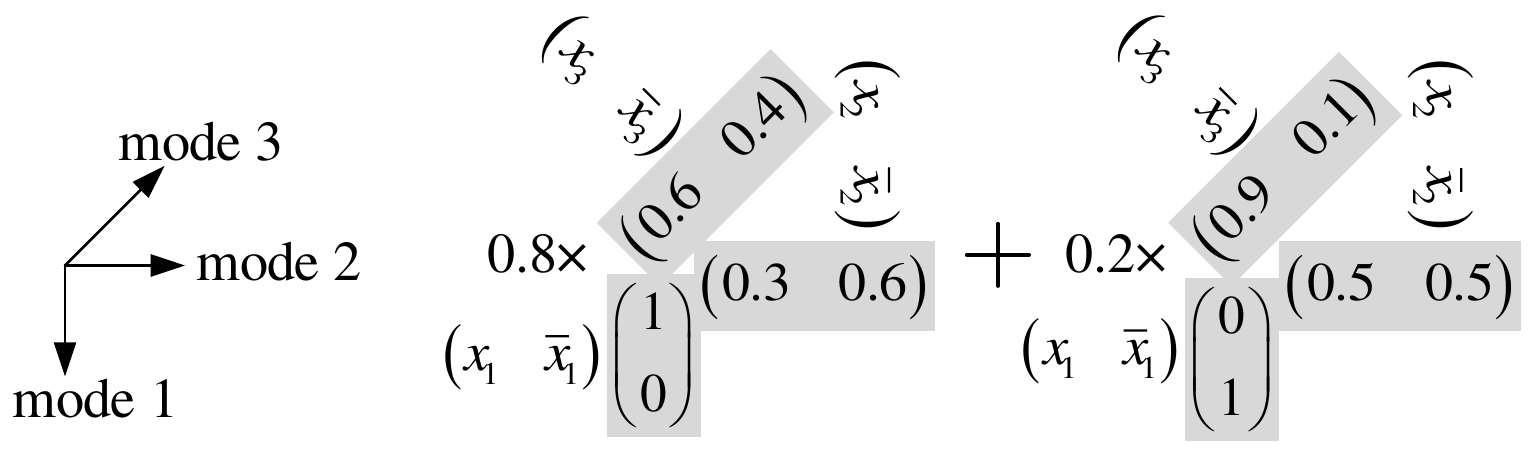}
\caption{The tSPN equivalence of the SPN in Fig.~\ref{fig:spn_ex}(b), where the two terms correspond to the two induced trees.}
\label{fig:tspn_ex}
\end{center}
\end{figure}
\begin{figure*}[t] 
\begin{center} 
\includegraphics[width=0.8\textwidth]{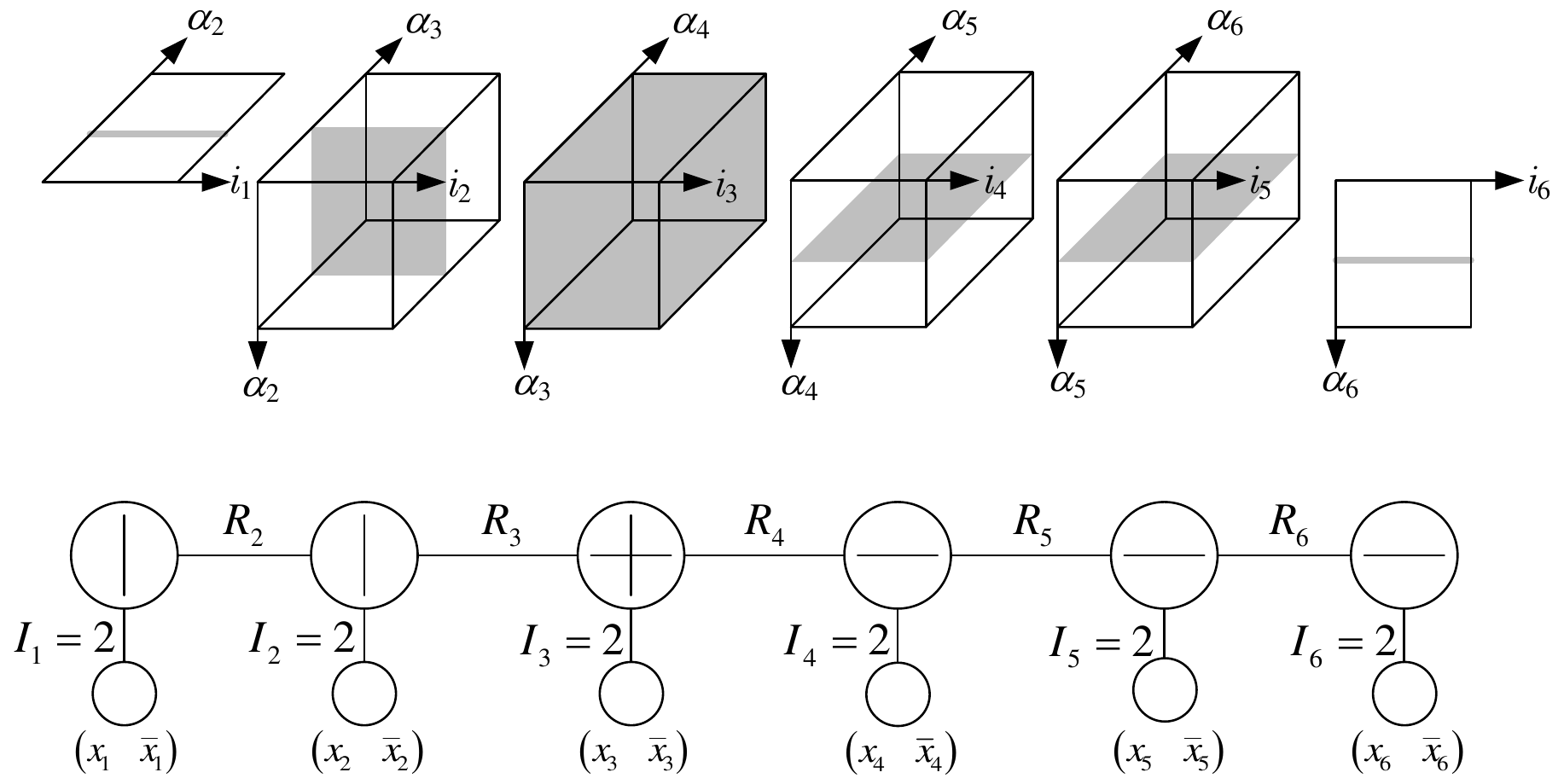}
\caption{The normalized tSPN corresponding to a normalized SPN, wherein the shaded parts within a core have entries summing up to unity. The vertical, cross, and horizontal lines in the lower tensor diagram denote left normalized, mixed and right normalized cores.}
\label{fig:normtspn}
\end{center}
\end{figure*}
The key motivation of this work comes from the important observation that an SPN induced tree  terminated at leaf nodes is in fact a rank-1 tensor. Using Fig.~\ref{fig:spn_ex}(b) again as an example, there are two such induced trees that can be regarded as the addition of two rank-1 tensor terms with mode products $(x_k ~\bar{x}_k)$ onto their $k$th mode, as shown in Fig.~\ref{fig:tspn_ex}. Consequently, summing all the rank-1 terms (gray shaded terms in Fig.~\ref{fig:tspn_ex}) produces a $d$-way cubical tensor of dimension $I=2$. This tensor can then hopefully be sufficiently approximated by a low-rank TT as a particular kind of tSPN. We aim at building a tSPN based on the TT structure, depicted in Fig.~\ref{fig:normtspn}, that satisfies the following constraints:
\begin{itemize}
\item The TT-cores have non-negative entries.
\item There is a \emph{mixed core}, whose position is arbitrary, with all its entries summing up to 1.
\item Every core to the left of the mixed core is a \emph{left normalized core}, which means that each of its vertical slices \mbox{$\ten{F}^{(k)}(:,:,\alpha_{k+1})$} sums up to 1.
\item Every core to the right of the mixed core is a \emph{right normalized core}, which means that each slice \mbox{$\ten{F}^{(k)}(\alpha_{k},:,:)$} sums up to 1.
\item When we encounter a slice that contains all zeros, then it means the two slices (one vertical and one horizontal) in two adjacent cores corresponding to the same $\alpha_k$ can be removed and the dimension $R_k$ is shrunk by one.
\end{itemize}
The first constraint ensures that a tSPN has non-negative weights. The remaining constraints ensure that the partition function $Z=1$ when $x_k=\bar{x}_k=1$ for all $k$'s. A tSPN obeying the above constraints is called a normalized tSPN in analogy to a normalized SPN. The left/right normalized cores and the mixed core are strongly analogous to the mixed-canonical form of a TT which consists of left/right orthogonalized cores and a mixed core. We remark that a tSPN having a TT structure automatically enforces the desired weight parameter sharing as well as a deep network. This is because each scalar (viz. probability) evaluation of a TT-based tSPN, when contracted with $(x_k~\bar{x}_k)$ onto its $k$th mode, $k=1,\ldots,d$, results in a matrix product using all TT-cores (cf. Definition~\ref{def:tspncontract}).

%\textcolor{red}{From Dr. Wong:\\
%I suggest we remove or rephrase the following 2 paragraphs, not to mention MNIST as we're waiting for Cong's results. Let's do this:
%\begin{enumerate}
%\item Mention we can convert from full tensor (cf. Fig. 5) to e.g., a TT based tSPN, but the conversion requires non-negative tensor factorization (cite something here) which is computationally expensive, and usually gives rise to full-rank TT which defeat the purpose of a compact tSPN representation;
%\item To overcome the issues above, we make use of the recently proposed TT identification approach, utilizing both samples and non-samples; 
%\item The above method allows us to formulate a low-rank TT representation of the full tensor, such a tSPN corresponds to a deep SPN.
%\item Reviewers may think we're approximating an approximation and may lead to crude and inaccurate probability distribution. We defend here by fore-telling that numerical experiments validate this approach and from the experimental plots we can see almost exact match. This can be attributed to our use of the exact training samples, and even more so, we utilize negative samples (via non-samples) too.
%\item Present the maths and algo box;
%\item To claim near-exact-match, we need to get the KL working to quantify our claim later on in experiments.
%\end{enumerate}
%}
Recalling from Fig.~\ref{fig:tspn_ex}, a tSPN is fully captured by a $d$-way cubical tensor $\ten{F}$ through summing all rank-1 terms (induced trees) extracted from the learned SPN. The conversion of such a full-tensor tSPN into a TT-based tSPN then boils down to converting $\ten{F}$ into its TT format $\ten{F}^{(1)},\ldots,\ten{F}^{(d)}$. 

A direct way to obtain the TT form of $\ten{F}$ is by regarding each rank-1 tensor term, corresponding to an induced tree, as a rank-1 TT and sum them all up into a new TT~\cite{oseledets2011tensor}. However, in this case, the TT ranks $R_2,\ldots,R_d$ equal the number of rank-1 terms and are therefore impractically high. Although TT-rounding by singular value decomposition (SVD) between successive cores may reduce the TT-ranks, it will generally destroy the non-negativity of weights and result in cores with negative values. Similar issues arise when we use non-negative tensor factorization (NTF) algorithms~\cite{Kim2014} on $\ten{F}$ which also produce a large number of rank-1 tensor terms. In fact, constructing the full tensor $\ten{F}$ explicitly is computationally prohibitive when the number of variables $d$ go beyond 17 on our computers. This motivates us to develop an SPN-to-tSPN construction algorithm, called spn2tspn, through a recently proposed tensor-network based nonlinear system identification method~\cite{batselier2017tensor} elaborated below.

%representation of the original tensor Consequently, To convert the learned SPN to our proposed TT-based tSPN, an obvious solution is to construct a full tensor from the tSPN structure we showed in Fig.~\ref{fig:tspn_ex} and then utilize an alternating linear scheme to minimize the difference between the full tensor and our TT-based tSPN, while the non-negative constraint of TT-based tSPN entries is achieved by a sequential non-negative least square subroutines. However, this solution may suffer the following problems. Firstly, it is impossible to construct a full tensor explicitly when the feature size $d$ is very large. Secondly, it will commonly cost a large amount of computation and produce a full-rank TT, which defeat the purpose of a compact tSPN representation. 

 %In this case, the following advantages can be achieved. Firstly, a low-rank TT-based tSPN can be identified efficiently through an alternating linear scheme. Secondly, the number of parameters in the resulted TT-based tSPN grows linearly along with the increase of data feature number $d$. Thirdly, by employing the non-negative least square solution during the TT-cores updating procedure, the non-negative constrain in our optimization problem is well satisfied.
\subsection{Algorithm: spn2tspn}
Starting from a learned SPN trained by a given dataset, we compute the probabilities of a set of training input samples as well as non-samples through exact inference. This step has linear complexity due to the SPN nature, and generates a set of multi-input single-output (MISO) data used for the identification of the underlying TT. On the one hand, training samples are meaningful data (positive samples) used in the SPN learning and therefore correspond to higher probabilities. On the other hand, we randomly generate some non-samples (negative samples) outside the dataset and feed them into the SPN for their probabilities which are mostly close to zero. We then utilize these MISO data to identify a TT-based tSPN by adapting the approach in~\cite{batselier2017tensor}. 

Now having a set of $N$ training samples and non-samples together with their probabilities, the goal is to obtain a tensor $\ten{F}\in\mathbb{R}^{2\times\cdots\times 2}$ in a TT format such that the probability distribution inferred from this TT-based tSPN closely tracks that of the SPN. We first collect the $N$ column vectors $(x_k ~\bar{x}_k)^T$ into the matrix $\mat{S}^{(k)}\in\mathbb{R}^{2\times N}$ for $k=1,\ldots,d$. Next, we formulate the optimization problem
\begin{align}
\label{eqn:ude}
\min_{\ten{F}} \;||\mat{S}^T \;\textrm{vec}(\ten{F}) - \mat{y}||_2^2,
\end{align}
such that $\ten{F}$ is a TT with non-negative cores $\ten{F}^{(k)}$, and $\mat{S}^T\in\mathbb{R}^{N\times 2^d}$ is computed from
\begin{align}
\label{eqn:khatriS}
\mat{S} &= \mat{S}^{(d)}\odot\mat{S}^{(d-1)}\odot\ldots\odot\mat{S}^{(1)},
\end{align}
and $\mat{y}\in\mathbb{R}^{N\times 1}$ is the collection of probabilities of $N$ samples and non-samples.

Following from~\cite{batselier2017tensor},~(\ref{eqn:ude}) is broken into least-squares subproblems of small sizes solved by the alternating linear scheme (ALS). To obtain non-negative $\ten{F}^{(k)}$'s, we further impose a non-negative constraint into each subproblem which is then solved by the non-negative least-squares (NNLS) method~\cite{lawson1995solving} within each ALS iteration. This problem formulation also resembles that of a tensor completion work that employs a TT format~\cite{TTC} but without the non-negativity constraint in place. We therefore refer the reader to the paper for details. In short, to solve for $\ten{F}^{(k)}$, one solves the following least-squares subproblem by NNLS%
\begin{align}
\label{eq:le}
\mat{y}
&=
\begin{pmatrix}
\mat{a}_{>k,1}^T\otimes\mat{s}^{(k)T}_1\otimes\mat{a}_{<k,1}\\
\mat{a}_{>k,2}^T\otimes\mat{s}^{(k)T}_2\otimes\mat{a}_{<k,2}\\
\vdots\\
\mat{a}_{>k,N}^T\otimes\mat{s}^{(k)T}_N\otimes\mat{a}_{<k,N}
\end{pmatrix}
\textrm{vec}(\ten{F}^{(k)}),
\end{align}
where $\mat{s}^{(k)}_l\in\mathbb{R}^{I_k\times 1} (1\leq l \leq N)$ denotes the $l$th column of $\mat{S}^{(k)}$, whereas $\mat{a}_{<k,l}^T$ and $\mat{a}_{>k,l}$ are auxiliary notations defined as:
\begin{align*}
\mat{a}_{<k,l}^T &:= (\ten{F}^{(1)} \times_2 \mat{s}^{(1)T}_l)  \ldots (\ten{F}^{(k-1)} \times_2 \mat{s}^{(k-1)T}_l) \in\mathbb{R}^{R_k},\\
\mat{a}_{>k,l} &:= (\ten{F}^{(k+1)} \times_2 \mat{s}^{(k+1)T}_l ) \ldots (\ten{F}^{(d)} \times_2 \mat{s}^{(d)T}_l) \in\mathbb{R}^{R_{k+1}}.
\end{align*} 
The proposed algorithm is summarized in Algorithm~\ref{alg:spn2tspn}.

\begin{table*}[h]
 \small
\centering
\caption{SPN and tSPN information for various datasets.}
\label{tbl:datasetInfo}
\vspace{1ex}
\begin{tabular}{@{}lrrrrrrrr@{}}
Dataset name & \# variables & \# layers &\# leaf nodes & \# weights & \# parameters & \# parameters & reduction  & TV~~~\\
&  &  & &  & (SPN)~~~ & (tSPN)~~~ &  ($\times$)~~~ & distance\\
\midrule
NLTCS & $16$& $19$  & $1420$  & $405$  &  $3245$  & $134$ & $24$ & $0.0105$\\
MSNBC & $17$ & $21$  & $5962$  &  $2181$ & $14105$  & $47$ & $300$ & $0.0178$\\
KDDCup2K  & $65$ & $39$  & $7543$   & $836$  &  $15922$  & $138$ & $115$ & $-$\\
Plants  & $69$ & $23$  & $10087$  & $743$  &  $20917$  & $208$ & $100$ & $-$\\
Audio & $100$& $23$  & $32148$  & $752$  &  $65048$  & $273$ &  $238$ & $-$\\  
Jester & $100$&$21$  & $23366$  &  $441$ &  $47173$  & $261$ &  $180$ & $-$\\  
% Netflix & $100$&  Binary  &    \\ 
% Accidents & $111$&  Binary  &    \\ 
Retail & $125$& $25$  & $16746$  &  $954$ &  $34446$ & $382$ &  $90$ & $-$\\ 
Pumb-star  & $163$& $27$  & $25978$ &  $750$ &  $52706$ & $259$  &  $203$ & $-$\\ 
DNA  & $180$& $13$  & $8589$  & $75$  &  $17253$  & $183$ & $94$ & $-$\\ 
Kosarek  & $190$& $27$  & $24583$  &  $1052$ &  $50218$  & $427$ &  $117$ & $-$\\
MSWeb  & $294$& $35$  & $18307$  & $1111$  &  $37725$  & $584$ &  $64$ & $-$\\
% Book & $500$&  Binary  &    \\
EachMovie  & $500$& $27$  & $19123$  & $256$  &  $38502$  &  $588$  &  $65$ & $-$ \\
% WebKB & $839$&  Binary  &    \\
Reuters-52  & $889$& $27$  & $91969$  & $415$  &  $184353$  &  $1017$  &  $181$  & $-$\\
20Newsgroup  & $910$& $29$  & $271036$ &  $856$ &  $542928$  &  $953$  &  $569$   & $-$\\
% BBC & $1058$&  Binary  &    \\
Ad   & $1556$& $77$  & $34874$  & $246$  &  $69994$  &  $2186$  &  $32$   & $-$\\
 \end{tabular}
 \label{tb:datasetInfor}
\end{table*}

\begin{alg} SPN-to-tSPN Conversion (spn2tspn) \\
\label{alg:spn2tspn}
\textit{\textbf{Input}}: A trained SPN and $N$ training samples and non-samples.\\
\textit{\textbf{Output}}: A compressed TT-based tSPN.
\begin{algorithmic}[1]
\State Construct $d$ input matrices $\mat{S}^{(1)},\mat{S}^{(2)},\ldots,\mat{S}^{(d)}$ as described above
\State Obtain the probabilities of $N$ samples and non-samples through inference by the trained SPN
\State Collect probabilities to construct the vector $\mat{y}$ 
\State Randomly initialize non-negative TT-cores
\While {stopping criteria not satisfied } 
% \State Repeat the above loop for $k=d,\ldots,2$.
\For{k=1,\ldots,d-1}
\State $\textrm{vec}(\ten{F}^{(k)}) \leftarrow$ solve~\eqref{eq:le} using NNLS
\State $\mat{b} \leftarrow$ sum over the first and second indices of $\ten{F}^{(k)}$
\State Identify zero entries $I \leftarrow$ find($\mat{b}\neq0$)
\State $\ten{F}^{(k)} \leftarrow$ $\ten{F}^{(k)}(:,:,I)\,{\times_3}\,\text{diag}(1./\mat{b}(I))$
\State $\ten{F}^{(k+1)} \leftarrow$ $\ten{F}^{(k+1)}(I,:,:)\,{\times_1}\,\text{diag}(\mat{b}(I))$
\EndFor
\For{k=d,\ldots,2}
\State $\textrm{vec}(\ten{F}^{(k)}) \leftarrow$ solve~\eqref{eq:le} using NNLS
\State $\mat{b} \leftarrow$ sum over the second and third indices of $\ten{F}^{(k)}$
\State Identify zero entries $I \leftarrow$ find($\mat{b}\neq0$)
\State $\ten{F}^{(k)} \leftarrow$ $\ten{F}^{(k)}(I,:,:)\,{\times_1}\,\text{diag}(1./\mat{b}(I))$
\State $\ten{F}^{(k-1)} \leftarrow$ $\ten{F}^{(k-1)}(:,:,I)\,{\times_3}\,\text{diag}(\mat{b}(I))$
\EndFor
% \State Repeat the above loop for $k=1,\ldots,d-1$.
\EndWhile
\end{algorithmic}
\end{alg}

Once a tSPN is built, a purposedly reserved portion of the dataset input samples (not used in the SPN learning) are used as test inputs and their SPN probability outputs are used as test outputs to check the quality of the tSPN. As will be shown later in experiments, almost exact matches are observed between the probabilities derived from the learned SPN and its tSPN couterparts, with the latter having substantially fewer parameters.

We note that SPN operations such as inference and most probable explanation (MPE) etc. can be easily mapped to the tSPN scenario which, being not central to the theme and also due to space limitation, are put into the \textbf{Supplemental Material} of this paper.

\section{Experiments}
\label{sec:exp}
We use fifteen benchmarking datasets from publicly available website\footnote{https://github.com/arranger1044/spyn/tree/master/data} and learn their corresponding SPNs by using the methods proposed in~\cite{Vergari2015}. The Python implementation\footnote{https://github.com/arranger1044/spyn} of the algorithms are used throughout our experiments. We then apply Algorithm~\ref{alg:spn2tspn} to transform the SPNs into tSPNs, and the results are summarized in Table~\ref{tbl:datasetInfo}. All experiments were run on a desktop computer with an Intel i5 quad-core processor at 3.2GHz and 16GB RAM. Implementations of the proposed algorithm, together with necessary data files for reproducing experimental results will be made publicly available after the peer review. To measure the ``closeness'' of the probability distributions between an SPN and its tSPN counterpart, we use the total variation (TV) distance~\cite{gibbs2002choosing}. However, we stress that this metric (essentially an L1 norm) is only handy when the size of the problem is relatively small, say, \# variables $d\leq 17$. When there are more variables like $d\geq 18$, our desktop computer fails to construct a $2^d$ tensor and compute the TV distance. That said, to still evaluate the tSPN conversion, we plot the probability profiles of training and testing data both in an SPN and its corresponding tSPN to visually compare their similarity.%

\begin{figure}[t] 
\begin{center} 
\includegraphics[width=0.5\textwidth]{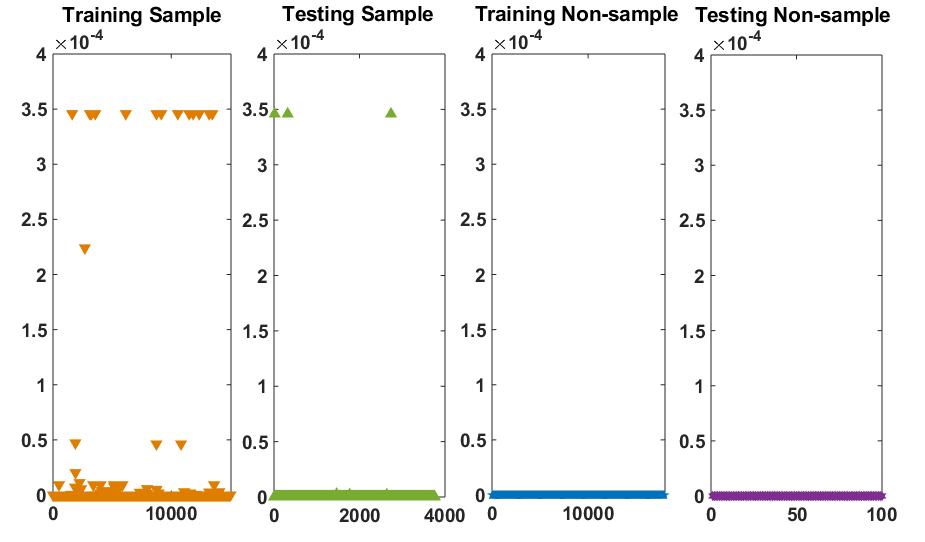}
\caption{Probability profiles of the learned SPN for \textit{20Newsgroup} dataset.}
\label{fig:c20ng_ori}
\end{center}
\end{figure}
\begin{figure}[t] 
\begin{center} 
\includegraphics[width=0.5\textwidth]{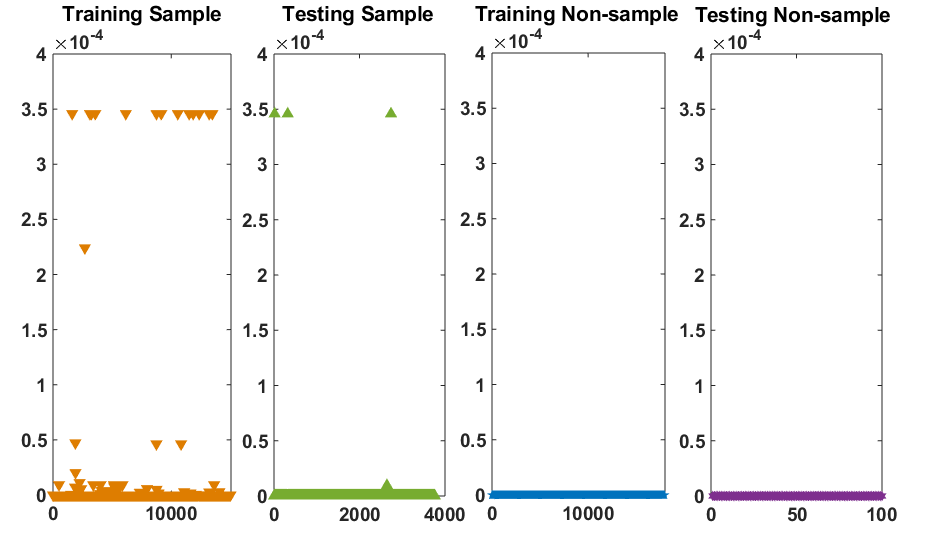}
\caption{Probability profiles of the constructed tSPN, at $569\times$ fewer parameters, for \textit{20Newsgroup} dataset.}
\label{fig:c20ng_new}
\end{center}
\end{figure}

As shown in Table~\ref{tbl:datasetInfo}, the number of parameters in tSPNs are significantly fewer than those in the original SPNs, while excellent match is still observed in their probability distributions as indicated by the TV distances and as depicted in Figs.~\ref{fig:c20ng_ori} and~\ref{fig:c20ng_new}. In the first two experiments, the number of variables are both no bigger than $17$, thus we are still able to compute the TV distances from reference SPNs to tSPNs. Among other experiments, we highlight that when converting the SPN of the \textit{20Newsgroup} dataset into a tSPN, a $569\times$ storage reduction is witnessed. Although explicitly computing $2^{910}$ probabilities for the TV distance evaluation is prohibitive, we can compare the probabilities of only parts of the samples and non-samples. In particular, in each of Figs.~\ref{fig:c20ng_ori} and~\ref{fig:c20ng_new}, we plot the probabilities of four sets of input samples, namely, training/testing samples/non-samples. Comparing the upper and lower figures, we observe essentially indistinguishable difference between the probability outputs of the SPN and tSPN, and similarly for other examples not plotted here. It is interesting to note that the testing samples/non-samples are never used in the spn2tspn conversion, but the resultant tSPN still manages to well capture the probabilities for these unseen data. This demonstrates the success of putting SPNs on tSPNs, which leads to substantial reduction in modeling parameters while having virtually no loss in accuracy.

Apparently, other tensor factorizations such as the CANDECOMP/PARAFAC canonical polyadic decomposition or Tucker decomposition~\cite{Kolda2009} may be used for tSPN, but according to our tests these are not as effective as the TT decomposition in preserving the depth and width semantics of SPNs. Furthermore, the identification algorithms for other tensor formats are not as simple and scalable as spn2tspn, and are therefore omitted. 

Backed by the experimental results, we highlight some additional advantages of TT-based tSPNs further to those already stated in the introduction:%
\begin{itemize}
	\item They often have small sizes via controlling the TT ranks, which translates into smaller network sizes and faster inference.
	\item Depth of a tSPN is inherently high (corresponding to the number of TT-cores), while its width (corresponding to TT-ranks) is usually set to be low. In terms of neural network language, this means higher expressive efficiency.
	\item The structural simplicity of a tSPN acts as a regularizer. In terms of neural network language, this implies a better inductive bias.
\end{itemize}
Of course, a natural question is whether data learning can be directly carried out on the tSPN structure, rather than going through the SPN learning followed by tSPN conversion. Research is underway along this direction and results will be reported in our upcoming work.

\section{Conclusions}
\label{sec:conclusion}
This paper is the first to propose a tensor implementation of an SPN called tSPN. An equivalence is drawn between an SPN with $d$ boolean variables to a $d$-way cubical tensor of dimension 2. The transformation of the latter into a tensor train (TT) representation then allows inherent sharing of originally distributed weights in an SPN tree, thereby leading to an often dramatic reduction of the number of network parameters as seen in various numerical experiments, at little or negligible loss of modeling accuracy. The TT-based tSPN also automatically guarantees a deep and narrow neural-network architecture. These promising new results have demonstrated tSPN to be a more natural form for realizing an SPN compared to its conventional tree structure.%

%\section*{Acknowledgment}

%Dr. Reveryrand would like to acknowledge the funding by XLIM, Limoges, France. 
%The authors would like to thank 

% or


% ============================================
%\appendices
%Appendix one text goes here %\cite{Roberg2010}.

%\section{}

%\section*{Acknowledgment}

%The authors would like to thank D. Root for the loan of the SWAP. The SWAP that can ONLY be usefull in Boulder...

\ifCLASSOPTIONcaptionsoff
  \newpage
\fi



%\begin{thebibliography}{1}


\bibliographystyle{IEEEtran}
\bibliography{IEEEabrv,tnnls_tspn}

\vfill


\end{document}